\documentclass[letterpaper, 10 pt, conference]{ieeeconf}  

\IEEEoverridecommandlockouts                              

\usepackage{graphicx} 
\usepackage{amsmath} 
\usepackage{amssymb}  
\usepackage{bm}

\usepackage{algorithm}
\usepackage{algpseudocode}
\usepackage{makecell}

\usepackage[backend=biber, language=english,
style=ieee,
sorting=none,      
giveninits=true,
maxbibnames=99,
maxcitenames=1,
maxnames=50,
block=none,
doi=false,
url=false,
isbn=false,
eprint=false]
{biblatex}
\usepackage{color}

\def\b0{\bm{0}}
\def\bd{\bm{d}}

\def\bp{\bm{p}}

\def\bD{\bm{D}}

\def\bQ{\bm{Q}}

\newcommand{\splitcell}[2][c]{%
  \begin{tabular}[c]{@{}c@{}}\strut#2\strut\end{tabular}%
}

\title{\LARGE \bf Enhanced 6D Pose Estimation for Robotic Fruit Picking*}

\author{Marco Costanzo, Marco De Simone, Sara Federico, Ciro Natale and Salvatore Pirozzi
\thanks{*This work was partially supported by the European Commission under the Horizon Europe research grant INTELLIMAN, project ID: 101070136.
This is a pre-print of a paper accepted for presentation @CODIT 2023.}
\thanks{Marco Costanzo, Marco De Simone, Sara Federico, Ciro Natale and Salvatore Pirozzi are with Dipartimento di Ingegneria,
        Università degli Studi della Campania Luigi Vanvitelli, 81031 Aversa (CE), Italy
        {\tt\small ciro.natale@unicampania.it}}%
}

\begin{document}

\maketitle
\thispagestyle{empty}
\pagestyle{empty}

\begin{abstract}
This paper proposes a novel method to refine the 6D pose estimation inferred by an instance-level deep neural network which processes a single RGB image and that has been trained on synthetic images only. The proposed optimization algorithm usefully exploits the depth measurement of a standard RGB-D camera to estimate the dimensions of the considered object, even though the network is trained on a single CAD model of the same object with given dimensions. The improved accuracy in the pose estimation allows a robot to grasp apples of various types and significantly different dimensions successfully; this was not possible using the standard pose estimation algorithm, except for the fruits with dimensions very close to those of the CAD drawing used in the training process. Grasping fresh fruits without damaging each item also demands a suitable grasp force control. A parallel gripper equipped with special force/tactile sensors is thus adopted to achieve safe grasps with the minimum force necessary to lift the fruits without any slippage and any deformation at the same time, with no knowledge of their weight.
\end{abstract}

\section{INTRODUCTION}

Having the ability to estimate the position and orientation of an object in space is a critical aspect for the autonomous execution of robotic manipulation tasks. This problem becomes very challenging if the objects of interest are natural objects, such as fruits or vegetables, due to the high variability of their shapes and dimensions.
Picking solutions in the grocery market able to handle many thousands of different items are starting to appear in the real world, such as the system first demonstrated by Ocado Group \cite{Ocado}. They usually adopt a suction cup to grasp the items from a bin and place them in another container. These kind of solutions might not be suitable to manipulate delicate fruits or vegetables, especially those with an irregular surface, such as strawberries. Limiting the damage to this kind of products is a primary goal of any automatic handling system. Therefore, as soon as grasping has to be performed using a multi-fingered device, the use of tactile sensing, possibly combined with soft grippers \cite{ClashHand}, seems to be a promising solution. Nevertheless, the first requirement to estimate the 6D pose of the item to be picked with an accuracy high enough to have a successful grasp is of paramount importance. As an additional requirement, the method for the pose estimation should be insensitive to the dimension of the item to be grasped since a significant variability is expected for fruits or vegetables.

\begin{figure}[ht]
\centering
\includegraphics[width=0.7\columnwidth]{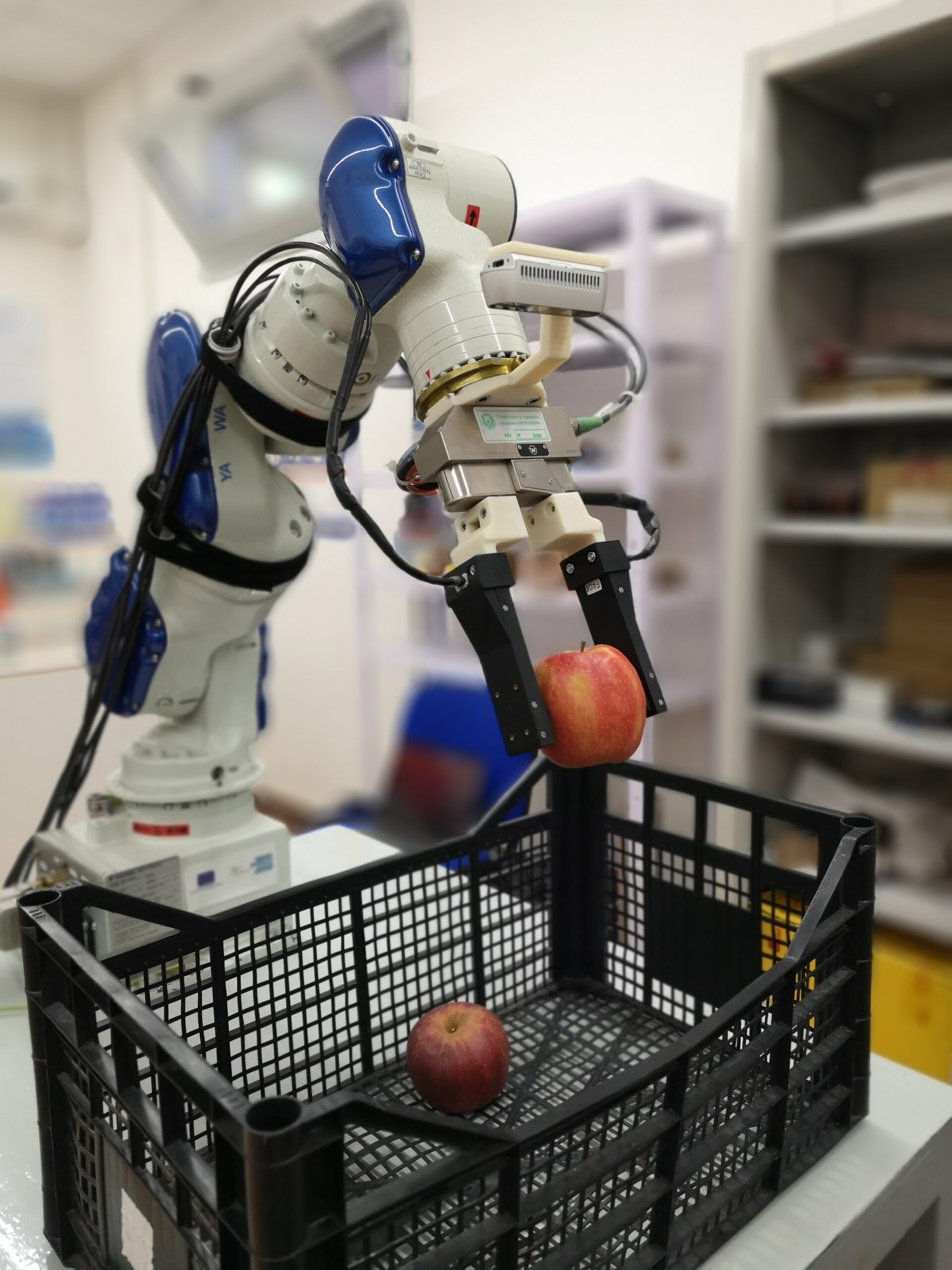}
\caption{Robot with sensorized gripper grasping an apple.}
\label{f:setup}       
\end{figure}

Most existing methods have focused on \textit{instance-level} object pose estimation \cite{DBLP:journals/corr/abs-1809-10790, DBLP:journals/corr/abs-1711-00199, DBLP:journals/corr/abs-1901-04780}, an approach that is ill-suited to fruit recognition, for which instead a \textit{category-level} algorithm \cite{DBLP:journals/corr/abs-2109-06161, DBLP:journals/corr/abs-1901-02970, DBLP:journals/corr/abs-2003-05848} would be more suited.
Modern instance-level methods can be categorized as template matching or regression techniques. Template matching techniques typically align known 3D CAD models to the observed 3D point clouds. Regression-based methods can estimate directly the 6D pose, as \cite{DBLP:journals/corr/abs-1711-00199} or predict 2D projected key-points, which are used by a PnP algorithm for solving the 6-DoF pose \cite{article}. All these techniques assume an exact 3D CAD model is available during training and they expect to work on the same instance object during inference, so they suffer from lack of scalability. In particular, the methods that use the PnP algorithm to retrieve the 6D pose, are particularly sensitive to the deviation of the target object from its CAD model.
Indeed, the most recent category-level methods do not require a CAD model of object instance for training and inference, thus promising a better solution for real world applications. Some of these try to address the problem of intra-class dimension variability by detecting the pose and the relative size of objects at inference time. However, to the best of our knowledge, these methods require a large dataset to be annotated by hand, that becomes very time consuming when 6DoF labels are required. Despite of this, the availability of a 3D CAD model can significantly alleviate the dataset creation thanks to its synthetic generation.
To merge the advantage of synthetic training data and the idea behind category-level pose estimation, we propose a post-refinement process that provides robustness to RGB instance-level pose estimation by adding the depth information. We address the problem for which, using only RGB data, we do not know if we are looking at a smaller but nearest instance of the object or at a bigger one but farther; such a problem is naturally reflected on its estimated pose. To overcome this issue, we propose an optimization algorithm, that, given the estimated pose of the target object, the depth map from a depth camera sensor, and the 3D CAD model used by the inference system, refines the estimated pose by estimating the real object dimensions. We assume that the pose estimation error is only caused by the different dimensions between the CAD model and the real object. Attracted by the possibility of realizing a synthetic dataset, as inference system, we adopt DOPE, the deep neural network proposed by \cite{DBLP:journals/corr/abs-1809-10790}. As suggest by the authors, in the dataset we merge a set of domain randomized and photorealistic data, genereted by the NVISII software \cite{DBLP:journals/corr/abs-2105-13962} to overcome the so called \textit{reality gap}. Our post-refinement method is located downstream of the EPnP used by DOPE, to improve the result of the neural network so as to make the estimated and refined pose sufficiently robust to grasp the object, independently from the fruit dimensions.

As explained before, estimating the pose of the fruit to be picked is not enough for a successful grasp that should avoid any damage to the object. Indeed, we adopt a parallel gripper equipped with the SUNTouch force/tactile sensors \cite{SENS2019}, which are able to measure the full 6D contact wrench. These measurements are then used in the slipping avoidance grasp control algorithm that automatically choose the lowest grasp force to lift the item without any knowledge of its weight \cite{COSTANZO2021}.

The experimental results of several picking tests of different types of apples with different dimensions show that the proposed robotic picking solution has a success rate much higher than the one ensured by the standard 6D pose estimation algorithm.
Fig. \ref{f:setup} shows the setup of the robotic picking solution.


\section{ENHANCED 6D POSE ESTIMATION}
\label{s:problem_formulation}
The \textit{instance-level} inference system adopted to detect the fruit in the scene and estimate its pose is DOPE, which has been proven effective in situations where the object to detect is a real instance of the CAD model used for training the neural network. Our aim is to make it suitable to estimate an accurate 6D pose for a class of apples, with dimensions that can largely deviate from those expected by the network.
DOPE recognizes the 6D pose of known and rigid objects with fixed dimensions by using only RGB data, and, in particular, by directly regressing the nine keypoints of a cuboid enclosing the recognized object, i.e., its eight vertices and centroid. The keypoints are then used by the PnP algorithm, that, given also the camera intrinsics and the CAD cuboid dimensions, is able to retrieve the 6D object pose.
It is obvious that this method is prone to positioning errors in case the object to detect is scaled with respect to the one used to train the network. This is clearly due to the lack of depth information. Indeed, DOPE will interpret the scale difference as a translation, nearest or farthest, of the target object. To overcome this limitation in case of an object with unknown dimensions, we aim at finding a method able to translate and scale an object without altering its representation in the image plane of the camera. To do this, we will exploit the depth map provided by a RGB-D camera.

\begin{figure}[ht]
\centering
\includegraphics[width=\columnwidth]{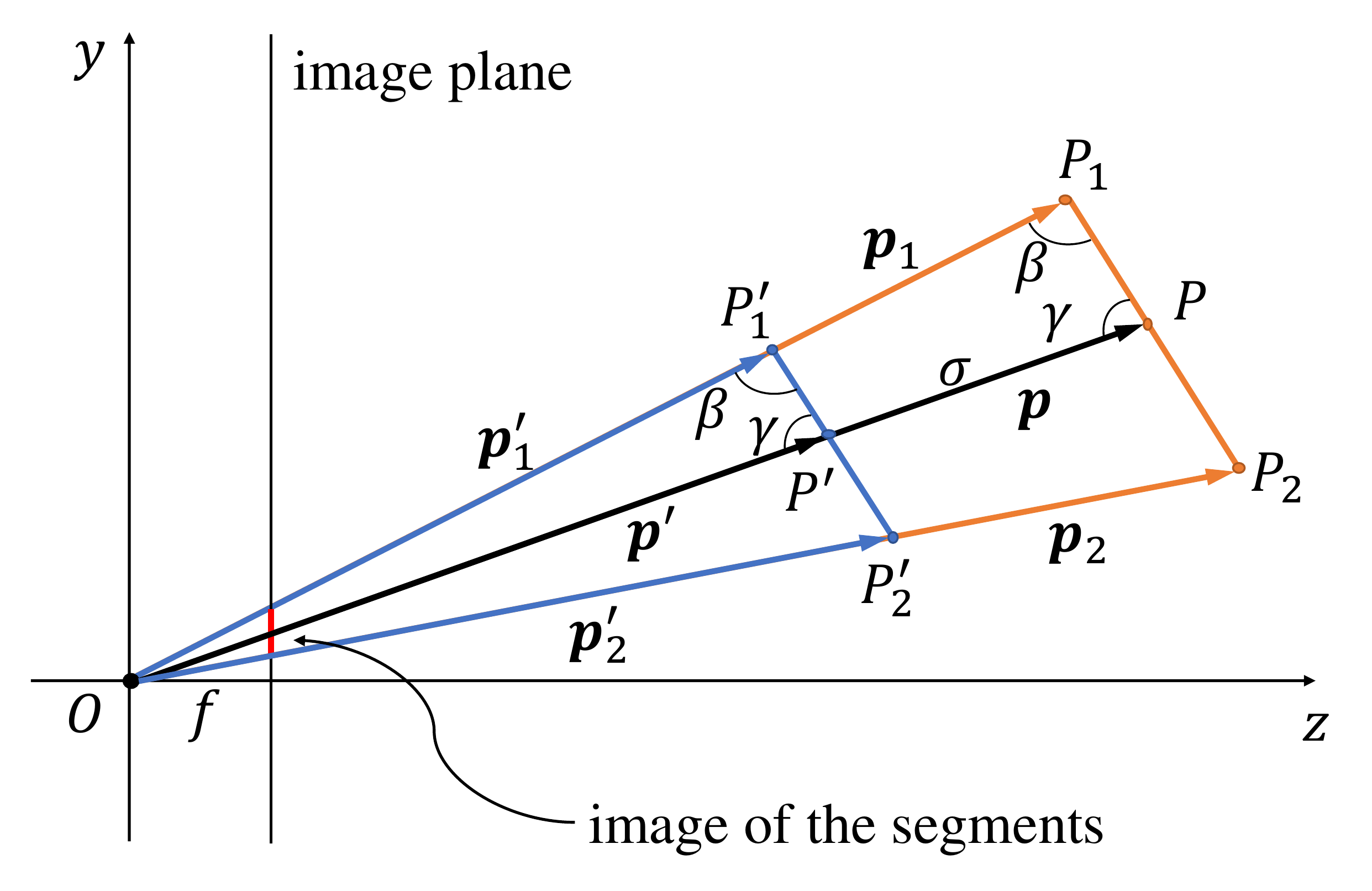}
\caption{Problem formulation: 2D approach.}
\label{f:segment-scaling}       
\end{figure}

Without loss of generality, with reference to Fig. \ref{f:segment-scaling}, consider a segment $S$ of length $s$ lying on the plane $y-z$ of the camera frame $O-xyz$, and a new segment $S'$ of length $s'$ obtained from both translation and scaling of $S$ by a scalar factor $\sigma$ along the direction identified by the vector $\bp$, representing a given point $P$ lying on $S$, in the aforementioned frame. This point $P$ represents the invariant point for scale.
Let $\bp'$ be the vector obtained by
\begin{equation}\label{eq:sigma-translation}
\bp' = \bp - {\sigma}{\frac{\bp}{\left\lVert \bp \right\rVert}}
\end{equation}
that identifies the new point $P'$ lying on $S'$.
The scaling factor $\mu(\sigma)$ is such that
\begin{equation}\label{eq:dimension generic relation}
s' = \mu(\sigma) s
\end{equation}
and the end points $P'_1$ and $P'_2$ of $S'$ belong to the straight lines with direction $\frac{\bp_1}{\left\lVert \bp_1 \right\rVert}$ and $\frac{\bp_2}{\left\lVert \bp_2 \right\rVert}$, respectively, with $\bp_1$ and $\bp_2$ identifying $S$ end points $P_1$ and $P_2$. In this way, given the camera frame $O-xyz$, with $z$ as the optical axis, the two segments $S$ and $S'$ are indistinguishable in the image plane, which is the plane parallel to $x-y$ camera plane located at distance $f$ from $O$ along $z$, being $f$ the camera focal length. 

Consider the two triangles $P_1OP$ and $P'_1OP'$; they are similar for obvious geometrical considerations, so we can state that
\begin{equation}\label{eq:triangles relation 1}
{\left\lVert \bp'_1 - \bp'\right\rVert} = {\frac{\left\lVert \bp' \right\rVert}{\left\lVert \bp \right\rVert}}{\left\lVert \bp_1 - \bp\right\rVert}
\end{equation}
Similarly, for the two triangles $P_2OP$ and $P'_2OP'$, it holds
\begin{equation}\label{eq:triangles relation 2}
{\left\lVert \bp'_2 - \bp'\right\rVert} = {\frac{\left\lVert \bp' \right\rVert}{\left\lVert \bp \right\rVert}}{\left\lVert \bp_2 - \bp\right\rVert}
\end{equation}

Combining \eqref{eq:triangles relation 1} and \eqref{eq:triangles relation 2}, we get the desired relation
\begin{equation}\label{eq:sigma scaling}
s' = \mu(\sigma) s, \-\hbox{with\ }\ \mu(\sigma) =  \frac{\left\lVert \bp' \right\rVert}{\left\lVert \bp \right\rVert},
\end{equation}
where $\mu(\sigma)$ denotes the sought scale factor depending on $\sigma$.
Our solution combines equations \eqref{eq:sigma-translation} and \eqref{eq:sigma scaling} to move and scale the CAD model in the virtual world without changing its appearance in the virtual camera image plane. In this way, generating for each $\sigma$ a different virtual depth map and comparing it with the real depth map captured by an RGB-D sensor, we can find the optimum $\sigma$ to overlap the CAD model to the real object.

Formally, let the estimated object pose by DOPE be represented by the vector $\hat\bp$ and the unit quaternion $\hat{\bm{Q}}$ indicating the position and the orientation of the recognized object with respect to the camera frame, respectively.

In order to estimate the optimum scalar factor $\sigma$ (and thus the scale factor $\mu$), the following procedure is proposed.


For any given $\sigma$, the CAD model is moved in the virtual world from the position $\hat\bp$ to the new position $\hat\bp'$ computed as in \eqref{eq:sigma-translation} while keeping the same orientation $\hat{\bm{Q}}$, and its nominal dimensions represented by the vector
\begin{align}
\bm{d}_{CAD}=\begin{bmatrix}d_x \ d_y \ d_z\end{bmatrix}^T
\end{align}
are scaled by the factor $\mu$ corresponding to the given $\sigma$. Then, the corresponding depth map is acquired, i.e.,
\begin{align}
\bm{\hat{D}}'(\sigma)=
\begin{bmatrix}
    \hat{d}_{11}'(\sigma) & \dots  & \hat{d}_{1w}'(\sigma)\\
    \vdots & \ddots & \vdots\\
    \hat{d}_{h1}'(\sigma) & \dots  & \hat{d}_{hw}'(\sigma)
\end{bmatrix}
\end{align}
where $w$ and $h$ are width and height number of pixels of the camera, respectively.
Let
\begin{align}
\bm{D}=
\begin{bmatrix}
    {d_{11}} & \dots  & {d}_{1w}\\
    \vdots & \ddots & \vdots\\
    {d}_{h1} & \dots  & {d}_{hw}
\end{bmatrix}
\end{align}
be the real depth values, then an optimization problem can be set based on the difference between $\bD$ and $\hat\bD'(\sigma)$, i.e.,
\begin{equation}\label{eq:f_obj}
\min_\sigma \frac{1}{\rho}\sum_{(i,j)\in \mathcal{P}} {({d_{ij}} - {\hat{d}_{ij}'{(\sigma)}})}^{2},
\end{equation}
where $\mathcal{P}$ is the set of pixels of the CAD model in the depth map with cardinality $\rho$. Note that $\sigma$ both translates and scales the CAD model in the virtual world, so that, for different $\sigma$ values, the same image is captured by the RGB sensor but different values are measured in the virtual depth map.

The optimal value $\sigma_{opt}$ is obtained by solving \eqref{eq:f_obj} using a standard deterministic constrained optimization algorithm with bounds $[-0.8\hat{p}_z, 0.8\hat{p}_z]$ and it is used to update the estimated position $\hat\bp$ according to \eqref{eq:sigma-translation} as well as to estimate the real object dimensions according to the optimal scale factor $\mu_{opt}$ as in \eqref{eq:sigma scaling}, i.e.,
\begin{equation}\label{eq:position-update}
\hat\bp'_{opt} = \hat\bp - {\sigma_{opt}}{\frac{\hat\bp}{\left\lVert \hat\bp\right\rVert}}
\end{equation}
\begin{equation}\label{eq:scale-update}
\bd' = \mu_{opt}\bd_{CAD},\;\hbox{with\ }\mu_{opt}= {\frac{\left\lVert\hat\bp'_{opt}\right\rVert}{\left\lVert \hat\bp\right\rVert}}
\end{equation}

\begin{figure}[t]
\centering
\includegraphics[width=0.7\columnwidth]{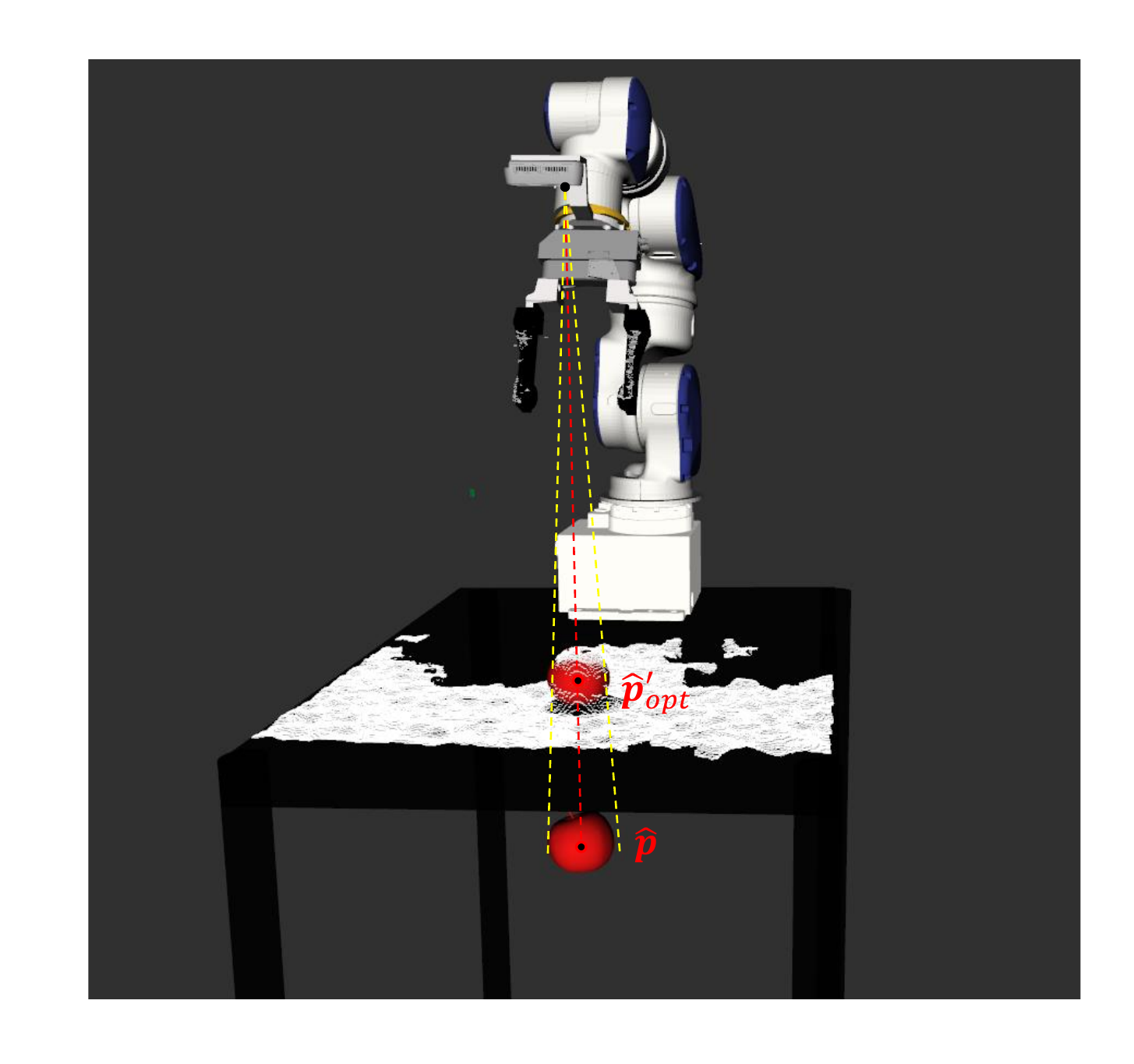}
\caption{Pose recognition of an apple with dimensions smaller then those of the CAD model. In white the real depth map coming from the RGB-D sensor.}
\label{f:dope_vs_our}       
\end{figure}

Fig. \ref{f:dope_vs_our} shows the CAD model placed in the pose estimated by DOPE and the scaled one in the refined pose.

Due to the variability of shapes and dimensions of the natural objects, it is not sure there is a perfect match between the CAD pixels in the virtual depth map and the target object pixels in the real depth map. Moreover, depending on the camera view, the set of real depth values used to solve the optimization problem  \eqref{eq:f_obj} can be altered by a possible occlusion of the object of interest. The corresponding pixels cannot be used in the post refinement process and they must be considered as outliers. For these reasons, we adopt the RANSAC algorithm \cite{10.1145/358669.358692} that robustly identifies the outliers through a linear regression.

\section{ROBOTIC FRUIT PICKING}\label{s:control}

\begin{figure}[t]
\centering
\includegraphics[width=\columnwidth]{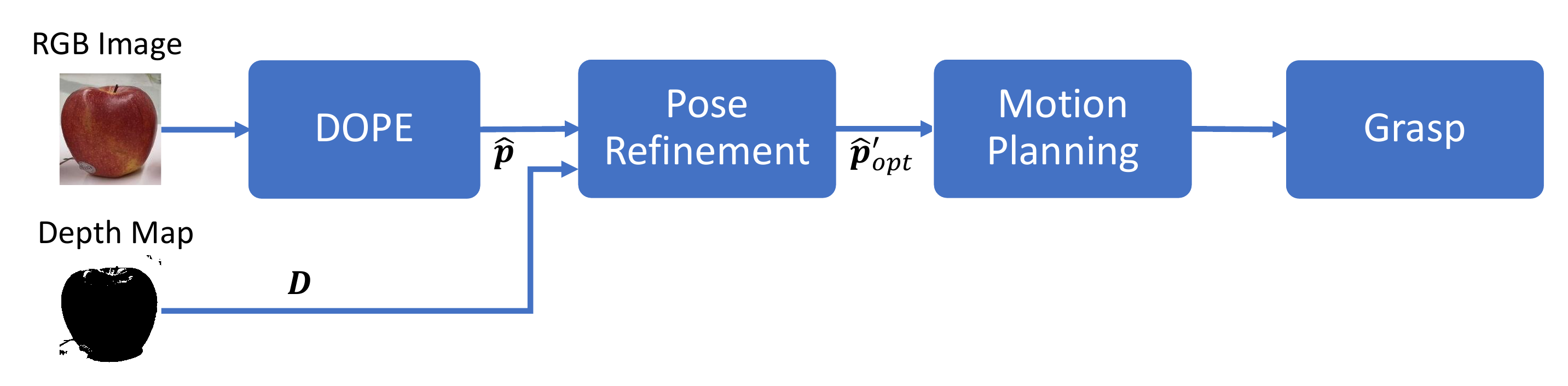}
\caption{Pipeline of the robotic fruit picking task execution.}
\label{f:pipeline}       
\end{figure}

\begin{figure}[t]
\centering
\includegraphics[width=\columnwidth]{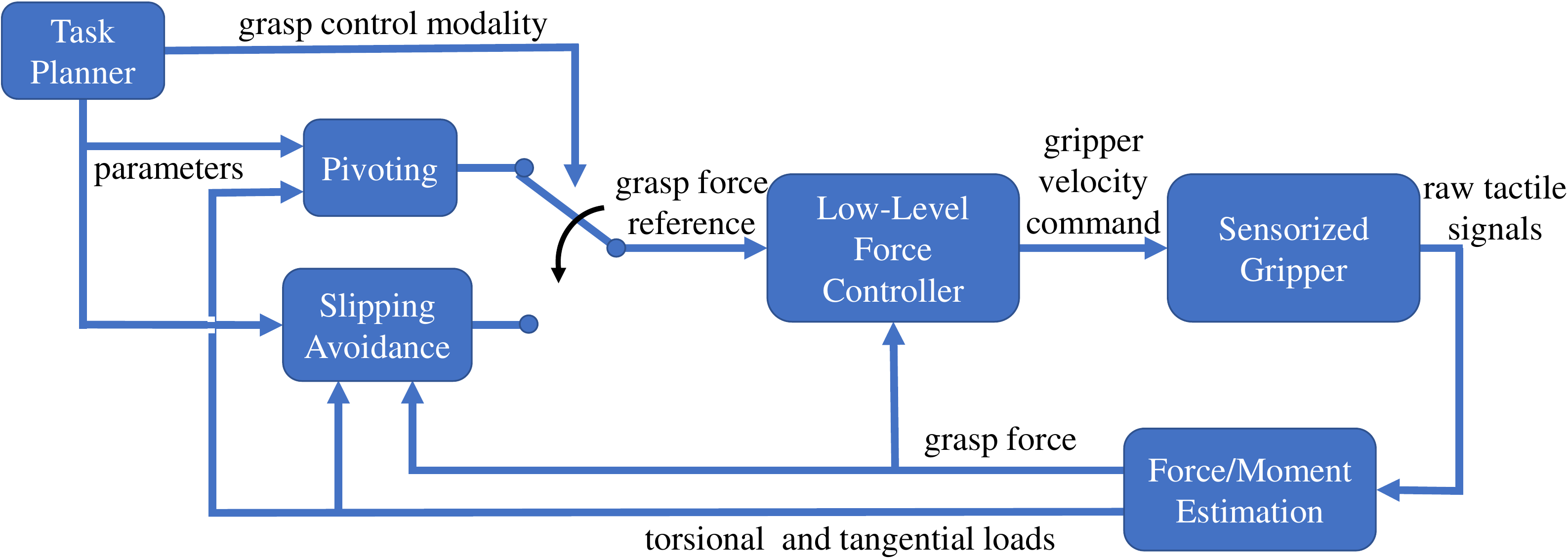}
\caption{Block scheme of the grasp controller. \textit{Slipping Avoidance}: eqs. (\ref{eq:slip_avoidance_control_law})-(\ref{eq:fnd}).}
\label{f:grasp_control_scheme}       
\end{figure}

The overall pipeline of the robotic fruit picking is reported in Fig. \ref{f:pipeline}, where the main steps include: DOPE inference, Pose refinement, Motion planning, Picking execution. The following subsections detail the robot control algorithms, the one devoted to automatically adjust the grasp force to properly grasp the fruit without damaging it and the one devoted to control the robot motion needed to bring the gripper to the grasp pose suitable to pick the item while avoiding collisions with other objects in the scene.

\subsection{Grasp Control}
\label{s:grasp}

Grasping the fruit with the lowest force compatible with its lifting without any slippage is a challenging objective and is here achieved by resorting to a model-based grasp controller whose details can be found in \cite{COSTANZO2021} and here only briefly recalled for the readers' convenience. The controller has two modalities, the \textit{slipping avoidance} mode and the \textit{pivoting} mode (see Fig. \ref{f:grasp_control_scheme}). The algorithm of each modality computes a grasp force reference that is then tracked by a low-level force controller by acting on the velocity of the gripper fingers. Only the slipping avoidance mode is used in this paper, and thus it is the only one summarized here.

In the \textit{slipping avoidance} modality, the grasp force reference is computed as the sum of two contributions, i.e.,
\begin{equation}\label{eq:slip_avoidance_control_law}
    f_n = f_{n_s} + f_{n_d},
\end{equation}
where $f_{n_s}$ and $f_{n_d}$ are the static and dynamic grasp force contribution, respectively. The algorithm to compute $f_{n_s}$ exploits the  tangential $f_t$ and torsional $\tau_n$ loads measured by the force/tactile sensors and computes the minimum force required to avoid slippage. It is based on the Limit Surface theory \cite{Cutkosky96} which generalizes the Coulomb friction law to the case of coupled linear and torsional frictional loads. With this method, the motion of the object grasped by two fingers is modeled as a planar slider, i.e., a rigid object subject to an instantaneous rotational motion about the instantaneous Center of Rotation (CoR) with angular velocity $\omega$.

The static contribution alone can avoid slippage only in quasi-static conditions, because in case of time-varying loads the maximum friction at the contact decreases as the rate of variation of the load increases \cite{Richardson76}, and such an effect is captured by the LuGre dynamic friction model \cite{Automatica2020}. Combining such a model with the Limit Surface theory it is possible to devise a slipping velocity observer to estimate $\omega$ during the slipping avoidance control phase, that is used while lifting the object. Since the weight is totally unknown and the inertial forces can be rapidly varying, the dynamic contribution $f_{n_d}$ is needed to avoid slippage; it is computed by means of the estimated velocity $\widehat\omega$ as
\begin{equation}\label{eq:fnd}
    f_{n_d} = \left|C_d\widehat\omega\right|,
\end{equation}
where $C_d$ is a suitable linear differential operator and the absolute value is needed to ensure that $f_{n_d} \ge 0$. More details about the slipping avoidance design and the stability of the closed-loop system are available in \cite{Automatica2020}.

\subsection{Motion Planning}
\label{s:motionplanning}
The robot motion planning is entirely realized in a ROS framework with the support of the MoveIt! platform. The latter became necessary to improve the autonomy of the robot during the picking task in a static environment. Thanks to the Occupancy Map Updater, through which the 3D perception is handled in MoveIt!, given in input a point cloud, the robot reconstructs the octomap of the surrounding environment. Through the Motion Planning plugin, we are able to plan a feasible trajectory that allows the robot to move to a suitable pre-grasp pose, avoiding the collision objects in the scene. As motion planning algorithm we adopt RRT*, powered by the default OMPL of MoveIt!.
To increase the chances to pick the target object although the obstacles in the scene, we adopt a grasping strategy based on the sampling of a sphere whose center is located in the refined position of the object.
Therefore, with respect to the world frame, the positions of the pre-grasp attempts are calculated as
\begin{equation}\label{eq:pre-grasp position attempt}
\bp_{att} = \hat{\bp}'_{opt} + r_o
\begin{bmatrix}
    \sin{\theta}\,{\sin{\alpha}}\\
    \sin{\theta}\,{\cos{\alpha}}\\
    \cos{\theta}
\end{bmatrix}
\end{equation}
where $\alpha$ is the rotation angle about the $z$ axis of the world frame, $\theta$ is the rotation angle about the gripper sliding vector and $r_o$ is the radius of the sphere.
Instead, each orientation is obtained as the quaternion
\begin{equation}\label{eq:pre-grasp orientation attempt}
\bQ_{e}^w = \Tilde{\bQ}\ast\bQ_z(\alpha)\ast\bQ_y(\theta),
\end{equation}
where $\ast$ denotes the quaternion product; $\bQ_z(\alpha)$ is the quaternion representing the elementary rotation about the $z$ axis of an angle $\alpha$, $\bQ_y(\theta)$ is the quaternion representing the elementary rotation about the $y$ axis of an angle $\theta$, and $\Tilde{\bQ}$ represents the rotation necessary to align the approach vector of the end-effector frame with the $z$ axis of the world frame.
The resulting $\bQ_{e}^w$ represents the end-effector orientation with respect to the world frame.

\section{EXPERIMENTAL RESULTS}
\label{s:experiments}

\begin{figure}[ht]
\centering
\includegraphics[width=\columnwidth]{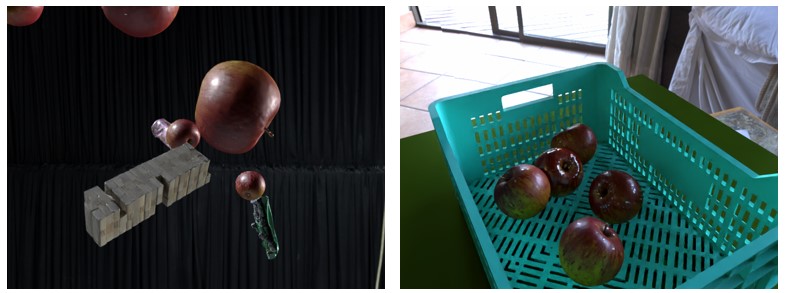}
\caption{Sample randomized (left) and photorealistic (right) domain data.}
\label{f:syntheticdata}       
\end{figure}

The objective of the experimental tests is to establish whether enhanced 6D pose estimation method is accurate enough to successfully realize robotic grasp of apples with different shapes and sizes.
The experiments are carried out on the 7-axis robot Yaskawa Motoman SIA5F, equipped with a WSG32 gripper by Weiss Robotics, sensorized with the SUNTouch force/tactile fingers \cite{SENS2019}, that provide a $5 \times 5$ tactile map and are calibrated to estimate the full contact wrench at $500\,$Hz between $0.5$-$8$\,N with an accuracy of $0.2$\,N.
The gripper accepts finger velocity commands updated at $50$\,Hz to control the grasp force.
As vision system we used a Realsense D435i camera mounted on the gripper.
The CAD model of the apple used to train DOPE is enclosed in a cuboid with dimensions in meters $\bd_{CAD}=\begin{bmatrix}0.092 \ 0.08 \ 0.092\end{bmatrix}^T$, where, according to the orientation of the CAD frame, $d_y=0.08\,$m represents the apple height.
The dataset consists of $120k$ frames, of which $100k$ frames are domain randomized and $20k$ are photorealistic (two sample images are reported in Fig. \ref{f:syntheticdata}). To improve the learning of the network, we randomly added from $5$ to $10$ instances of the CAD model and $5$ distractors in each frame.
The network was trained for $120$ epochs with a batch size of $116$ and a learning rate of $0.0001$. For both training and testing we used a Dell Precision 5820 workstation with two GPUs Quadro RTX 6000.

\begin{figure}[ht]
\centering
\includegraphics[width=\columnwidth]{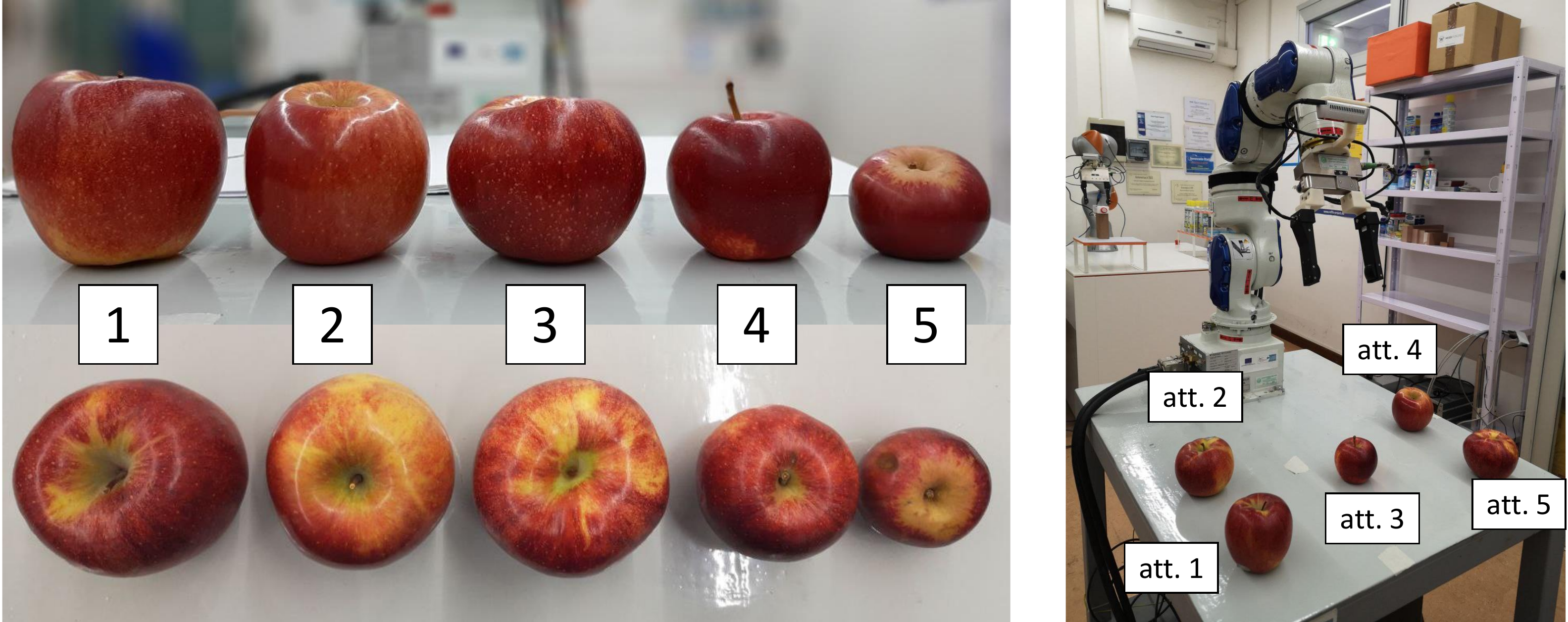}
\caption{On the left the five apples used for the experiments; on the right the five locations for grasp attempts}
\label{f:objectset}       
\end{figure}

To obtain the quantitative results, we used five different typologies of apples (see Fig. \ref{f:objectset}), three of them have dimensions similar to those of the CAD model and two have dimensions significantly different, as shown in Tab.~\ref{tab:apple dimensions}. We put them at five different locations on the table on which the robot is fixed to investigate the grasp success rate with and without our 6D pose refinement method.

\begin{table}[t]
\caption{Test with pose refinement in att. 3}
\label{tab:att3_with_correction}
\begin{center}
\begin{tabular}{|c||c|c|c|c|c|c|}
\hline
& \splitcell{Refined \\ Pos. [m]} & \splitcell{Centroid\\ Err. [m]} & \splitcell{Est. Obj.\\ Dim. [m]} & \splitcell{Dim.\\ Err. [m]} & $\mu_{opt}$\\
\hline
Apple 1 & $\begin{array}{r}
     -0.026 \\
     -0.525 \\
     0.045 \\
\end{array}$ & $-0.009$ & $\begin{array}{c}
0.091\\
0.079\\
0.089\\
\end{array}$ & $0.002$ & $0.928$\\
\hline
Apple 2 & $\begin{array}{r}
     -0.042 \\
     -0.519 \\
     0.041 \\
\end{array}$ & $-0.001$ & $\begin{array}{c}
0.087\\
0.077\\
0.086\\
\end{array}$ & $0.005$ & $0.897$\\
\hline
Apple 3 & $\begin{array}{r}
     -0.035 \\
     -0.521 \\
     0.031 \\
\end{array}$ & $0.0058$ & $\begin{array}{c}
0.097\\
0.084\\
0.095\\
\end{array}$ & $0.002$ & $0.995$\\
\hline
Apple 4 & $\begin{array}{r}
     -0.028 \\
     -0.520 \\
     0.024 \\
\end{array}$ & $0.0014$ & $\begin{array}{c}
0.074\\
0.065\\
0.073\\
\end{array}$ & $0.008$ & $0.763$\\
\hline
Apple 5 & $\begin{array}{r}
     -0.031 \\
     -0.526 \\
     0.011 \\
\end{array}$ & $0.012$ & $\begin{array}{c}
0.063\\
0.055\\
0.062\\
\end{array}$ & $0.0081$ & $0.648$\\
\hline
\end{tabular}
\end{center}
\end{table}

\begin{table}[t]
\caption{Test without pose refinement in att. 3}
\label{tab:att3_no_correction}
\begin{center}
\begin{tabular}{|c||c|c|c|}
\hline
& Estimated Position [m] & Centroid Error [m] \\
\hline
Apple 1 & $\begin{array}{r}
     -0.026\\
     -0.532 \\
     0.002 \\
\end{array}$ & $0.038$ \\
\hline
Apple 2 & $\begin{array}{r}
     -0.045 \\
     -0.535 \\
     -0.029 \\
\end{array}$ & $0.013$  \\
\hline
Apple 3 & $\begin{array}{r}
     -0.035 \\
     -0.522 \\
     0.029 \\
\end{array}$ & $0.007$ \\
\hline
Apple 4 & $\begin{array}{r}
     -0.034 \\
     -0.564 \\
     -0.172\\
\end{array}$ & $0.203$\\
\hline
Apple 5 & $\begin{array}{r}
     -0.028 \\
     -0.617 \\
     -0.349 \\
\end{array}$ & $0.372$\\
\hline
\end{tabular}
\end{center}
\end{table}

\begin{table}[t]
\caption{Apples dimensions to test the pipeline}
\label{tab:apple dimensions}
\begin{center}
\begin{tabular}{|c||c|c|c|}
\hline
& $d_x$ [m] & $d_y$ [m] & $d_z$ [m]\\
\hline
Apple CAD & $0.092$ & $0.081$ & $0.092$\\
\hline
Apple 1 & $0.089$ & $0.081$ & $0.089$\\
\hline
Apple 2 & $0.091$ & $0.085$ & $0.091$\\
\hline
Apple 3 & $0.093$ & $0.073$ & $0.093$\\
\hline
Apple 4 & $0.073$ & $0.064$ & $0.073$\\
\hline
Apple 5 & $0.062$ & $0.047$ & $0.062$\\
\hline
\end{tabular}
\end{center}
\end{table}

We express all the estimated poses in a reference frame with the origin located on the table surface and the $z$ axis pointing in the upward direction. In this way, we can evaluate the $z$ ground truth by simply halving the height ($d_y$) of each apple.
In addition to the grasp result, we report two metrics useful to evaluate the accuracy of our pose estimation method. The first one is the \emph{centroid error} defined as the difference between $d_y/2$ and the estimated centroid position along the $z$ direction because it is very simple to evaluate; the second one is the \emph{dimensional error} defined as the norm of the vector $\bd'_{opt}-\bd_{\rm Apple\ i}$, i =1,$\ldots$,5, where $\bd_{\rm Apple\ i}$ are the five vectors reported in Tab. \ref{tab:apple dimensions}.

Firstly, we evaluate the results obtained with each apple in att. 3 (with reference to Fig. \ref{f:objectset}). Table \ref{tab:att3_no_correction} shows how the error made by DOPE inference is mainly along the $z$ component of the estimated object position and, as expected, it becomes larger when the apple dimensions are significantly different from those expected by the neural network (e.g., for Apple 4 and Apple 5). Table \ref{tab:att3_with_correction} shows that the proposed pose refinement is able to correct this error with each kind of apple (as evident by evaluating the error with respect to the $z$ ground truth as above). The resulting pose is accurate enough to allow the robot to grasp the target apple with a higher success rate. Table \ref{tab:att3_with_correction} shows how the proposed method is able to estimate the  dimensions of the real fruits with an error in the millimeters range.

Then, a comparison of Tab. \ref{tab:grasp_rate_no_correction} and Tab. \ref{tab:grasp_rate_with_correction} shows how the pose refinement increases the grasping success rate, computed on $3$ repetitions of each grasp attempt, also when the dimensions of the apples are similar to the CAD model used for training (referring to Apple 1, Apple 2 and Apple 3 as defined in Tab. \ref{tab:apple dimensions}). In fact, the poses obtained by DOPE are not always accurate enough to ensure the success of the pick, due to the high sensibility even for small deviations from the CAD model used for training. Note how picking apples 4 and 5 is impossible by using the pose estimated by DOPE, due to the substantial difference in the dimensions of the real fruit with respect to the CAD model used in the training of the network.
The success rates obtained for Apple 4 and 5 using the refined pose are less than 100\% due to the failure of the motion planning caused by collisions between the fingers and the support table in view of the small size of the fruits, rather than the low accuracy in the pose estimation.


\begin{table}[t]
\caption{Grasping success rate without pose refinement}
\label{tab:grasp_rate_no_correction}
\begin{center}
\begin{tabular}{|c||c|c|c|c|c|}
\hline
& ATT. 1 & ATT. 2 & ATT. 3 & ATT. 4 & ATT. 5 \\
\hline
Apple 1 & $100\%$ & $66\%$ & $0\%$ & $100\%$ & $100\%$\\
\hline
Apple 2 & $33\%$ & $33\%$ & $33\%$ & $0\%$ & $0\%$\\
\hline
Apple 3 & $66\%$ & $33\%$ & $33\%$ & $33\%$ & $33\%$\\
\hline
Apple 4 & $0\%$ & $0\%$ & $0\%$ & $0\%$ & $0\%$ \\
\hline
Apple 5 & $0\%$ & $0\%$ & $0\%$ & $0\%$ & $0\%$ \\
\hline
\end{tabular}
\end{center}
\end{table}

\begin{table}[t]
\caption{Grasping success rate with pose refinement}
\label{tab:grasp_rate_with_correction}
\begin{center}
\begin{tabular}{|c||c|c|c|c|c|}
\hline
& ATT. 1 & ATT. 2 & ATT. 3 & ATT. 4 & ATT. 5 \\
\hline
Apple 1 & $100\%$ & $100\%$ & $100\%$ & $100\%$ & $100\%$\\
\hline
Apple 2 & $100\%$ & $100\%$ & $100\%$ & $100\%$ & $100\%$\\
\hline
Apple 3 & $66\%$ & $66\%$ & $100\%$ & $100\%$ & $100\%$\\
\hline
Apple 4 & $66\%$ & $100\%$ & $100\%$ & $100\%$ & $100\%$ \\
\hline
Apple 5 & $66\%$ & $33\%$ & $66\%$ & $66\%$ & $66\%$ \\
\hline
\end{tabular}
\end{center}
\end{table}

\section{CONCLUSIONS}

The paper presented a perception and control framework for robotic fruit picking. The perception algorithm starts from a state-of-the-art object pose estimation based on a deep neural network, that is then enhanced by an optimization algorithm which estimates the dimensions of the fruit by resorting to depth measurements. The actual grasp of the fruit is performed by using a grasp force controller exploiting force/tactile sensing, so as to ensure a safe grasp with a minimal force to avoid any kind of damage to the fresh fruit. Experimental results confirm that the new perception pipeline outperforms the state-of-the-art method by significantly increasing the number of successful grasps of various types of apples with relevant dimensional variability with respect to the fruit sample used for the neural network training based on synthetic data only.
Future work will be devoted to visual tracking of possibly moving fruits, e.g., on a conveyor belt, as well as to more sophisticated manipulation actions additional to the simple grasp, such as pushing maneuvers to enable more feasible grasps, e.g., in case the fruits are too close to each other or to the border of the container.










\printbibliography

\end{document}